\documentclass[runningheads]{llncs}
\newsavebox\CBox 
\def\textBF#1{\sbox\CBox{#1}\resizebox{\wd\CBox}{\ht\CBox}{\textbf{#1}}}
\usepackage{graphicx}
\usepackage{multirow}
\usepackage{amssymb}
\usepackage{booktabs}
\usepackage{hyperref}

\newcommand{\N}{\mathcal{N}}
\newcommand{\mha}{\mathrm{MHA}}
\newcommand{\Q}{\mathcal{Q}}
\newcommand{\X}{\mathbf{x}}
\newcommand{\R}{\mathbf{r}}

\usepackage{adjustbox}
\usepackage{array}
\usepackage{pifont}
\usepackage[table, dvipsnames]{xcolor}
\usepackage{colortbl}
\usepackage{nicematrix}
\usepackage{subcaption}
\usepackage{pict2e,picture}
 
\def\textBF#1{\sbox\CBox{#1}\resizebox{\wd\CBox}{\ht\CBox}{\textbf{#1}}}

\newcolumntype{R}[2]{%
    >{\adjustbox{angle=#1,lap=\width-(#2)}\bgroup}%
    l%
    <{\egroup}%
}
\newcommand{\xmark}{\ding{55}}%

\newcommand*\rot{\multicolumn{1}{R{45}{1em}}}

\makeatletter
\DeclareRobustCommand{\Arrow}[1][]{%
\check@mathfonts
\if\relax\detokenize{#1}\relax
\settowidth{\dimen@}{$\m@th\rightarrow$}%
\else
\setlength{\dimen@}{#1}%
\fi
\sbox\z@{\usefont{U}{lasy}{m}{n}\symbol{41}}%
\begin{picture}(\dimen@,\ht\z@)
\roundcap
\put(\dimexpr\dimen@-.7\wd\z@,0){\usebox\z@}
\put(0,\fontdimen22\textfont2){\line(1,0){\dimen@}}
\end{picture}%
}
\makeatother
\newcommand{\veryshortrightarrow}{\hspace{.2mm}\scalebox{.8}{\Arrow[.1cm]}\hspace{.2mm}}
\begin{document}
\newcommand{\cxmark}{\ding{51}\hspace{-1.75mm}\ding{55}}
\renewcommand*\rot{\multicolumn{1}{R{45}{1em}}}


\title{X-TRA: Improving Chest X-ray Tasks with Cross-Modal Retrieval Augmentation}
\titlerunning{X-TRA: Cross-Modal Retrieval Augmentation for Chest X-rays}
%
\author{Tom van Sonsbeek \and Marcel Worring}
%
%
\institute{University of Amsterdam\\
\email{\{t.j.vansonsbeek, m.worring\}@uva.nl}}
\maketitle              
\begin{abstract}

An important component of human analysis of medical images and their context is the ability to relate newly seen things to related instances in our memory. In this paper we mimic this ability by using multi-modal retrieval augmentation and apply it to several tasks in chest X-ray analysis. By retrieving similar images and/or radiology reports we expand and regularize the case at hand with additional knowledge, while maintaining factual knowledge consistency. The method consists of two components. First, vision and language modalities are aligned using a pre-trained CLIP model. To enforce that the retrieval focus will be on detailed disease-related content instead of global visual appearance it is fine-tuned using disease class information. Subsequently, we construct a non-parametric retrieval index, which reaches state-of-the-art retrieval levels. We use this index in our downstream tasks to augment image representations through multi-head attention for disease classification and report retrieval. We show that retrieval augmentation gives considerable improvements on these tasks. Our downstream report retrieval even shows to be competitive with dedicated report generation methods, paving the path for this method in medical imaging.

\end{abstract}
\section{Introduction}
The promise of automated deep learning systems to assist radiologists is enormous. At the moment, important milestones, such as better consistency or even better performance have been achieved on an increasing number of use-cases~\cite{zhou2021review,liu2019comparison}. A source of inspiration in further improvement of these efforts is the way humans register and analyze images, which for deep learning has shown to be effective in the past \cite{litjens2017survey,zhou2021review}. 

In any analysis, a doctor provides the memory and knowledge to place what is currently seen in the context of what has been seen before. In principle this can be compared to  what implicitly happens at scale in any deep learning method. A doctor's analysis is not implicit though. Their analysis process can be described and verified. We wonder whether (medical) deep learning methods could benefit from an explicit memory/knowledge infusion. 

Making deep learning methods more explicit in terms of using past observations has already been studied in Natural Language Processing (NLP), in the form of retrieval augmentation~\cite{lewis2020retrieval,pasupat2021controllable}. Supplementing data by retrieving relevant retrieved information can lead to performance gains \cite{gur2021cross}. This process can be thought of to work as both an enrichment and regularization process. A benefit of retrieval augmentation is that context from a trusted knowledge source is used as a supplement \cite{siriwardhana2022improving,komeili2022internet}. The versatility of retrieval augmentation, which essentially provides a non-parametric memory expansion, is gaining traction in the multi-modal field \cite{gur2021cross,ramos2022smallcap}. 

Multi-modal data modalities typically have different strengths leading to a strong and a weak data modality~\cite{zhou2021review}. For instance, radiology  reports generally contain richer and more complete information than X-rays, since the report is essentially a clinician's annotation~\cite{pooch2019can}. With retrieval augmentation information can be transferred explicitly from the strong to the weak modality. 

A reason retrieval augmentation methods are not yet adopted for medical applications lies in the weakness of retrieval methods for the medical domain. Retrieval in the general domain is focused on global image regions~\cite{li2018large,ionescu2022overview} whereas in medical images global features, such as body/organ structure are similar across patients. Meanwhile more fine-grained aspects are more discriminating as disease indicators, but are easily overlooked. The need for fine-grained results makes medical image retrieval magnitudes more complex.

We propose X-Ray Task Retrieval Augmentation (X-TRA), a framework for retrieval augmentation in a multi-modal medical setting, specifically designed for X-ray and radiology report analysis. To do so we introduce a cross-modal retrieval model and retrieval augmentation method. We make the following contributions. 

\begin{itemize}
    \item We propose a CLIP-based multi-modal retrieval framework with a dedicated fine-tuning component for efficient content alignment of medical information which improves state-of-the-art results in multi- and single- modal retrieval on radiology images and reports.
    \item We introduce a multi-modal retrieval augmentation component for disease classification and report retrieval pipelines. 
    \item We show that our method (1) reaches state-of-the-art performance both in multi-label disease classification and report retrieval. (2) Our report retrieval is competitive with dedicated report generation methodologies. (3) We show the cross-dataset versatility and the limitations of our method. 
\end{itemize}

\section{Related Work}
\paragraph{Multi-modal alignment}
The introduction of Transformers for natural language processing (NLP) accelerated the development of integrated vision-language (VL) alignment models suitable for various VL-tasks, such as ViLBERT~\cite{lu2019vilbert}, LXMERT~\cite{tan2019lxmert} and SimVLM~\cite{wang2021simvlm}. These methods provide alignment on region to sentence- or word- level scale. The next step in multi-modal alignment was made by methods using contrastive learning combined with substantially larger datasets. Examples are CLIP~\cite{radford2021learning} and ALIGN~\cite{jia2021scaling} which significantly outperform existing methods by using datasets for training consisting of 400M and 1.8B VL-pairs respectively. Domain-specific versions of CLIP, which is open-source, have been fine-tuned with additional data, such as PubMedCLIP \cite{EslamiDeMeloMeinel2021CLIPMedical}. 

\paragraph{Retrieval Augmentation}
The origin of retrieval augmentation lies in the NLP field. It was created to fully utilise the power of large datasets. With retrieval augmentation we are not only dependent on a parametric model, but can also supplement data as a non-parametric component. Previous methods have shown the simple yet effective and versatility working of retrieval augmentation in a number of applications ~\cite{komeili2022internet,guu2020retrieval,siriwardhana2022improving}.
\paragraph{Retrieval in Medical Imaging}
Up until recently the only retrieval methods in medical imaging were tailored hand-crafted methods \cite{li2018large}. With access to large datasets and pre-trained methods the balance shifted towards making automated retrieval methods \cite{qayyum2017medical,Hu_2022_WACV}. Especially in the histopathology and radiology domain major strides were made with retrieval methods~\cite{endo2021retrieval,ionescu2022overview}. The use of text to improve image retrieval has been adopted for improving chest X-ray retrieval. Yu \textit{et al.}\cite{yu2021multimodal} use CNN and word2vec features for multi-modal alignment and retrieval. Zhang \textit{et al.}~\cite{zhang2022category} approach this problem with a hash-based retrieval method. 

\paragraph{Retrieval for chest X-ray analysis}
Common tasks in chest X-ray analysis are disease classification and report generation \cite{johnson2019mimic,chambon2022roentgen,li2022self}. Using retrieval for report generation has been a common approach. The approaches often entail the use of retrieved information as an input or template for a decoder which crafts a custom report \cite{pino2021clinically,wang2022cross,yang2021writing}. Augmentation of chest X-ray tasks with synthetically generated diffusion-based images was shown to be possible~\cite{chambon2022roentgen}, however the clinical use of non-genuine images can lead to complications and is not undisputed~\cite{zhou2021review}. 

\section{Methods}
\begin{figure}[!t]
    \centering
    \includegraphics[width=\linewidth]{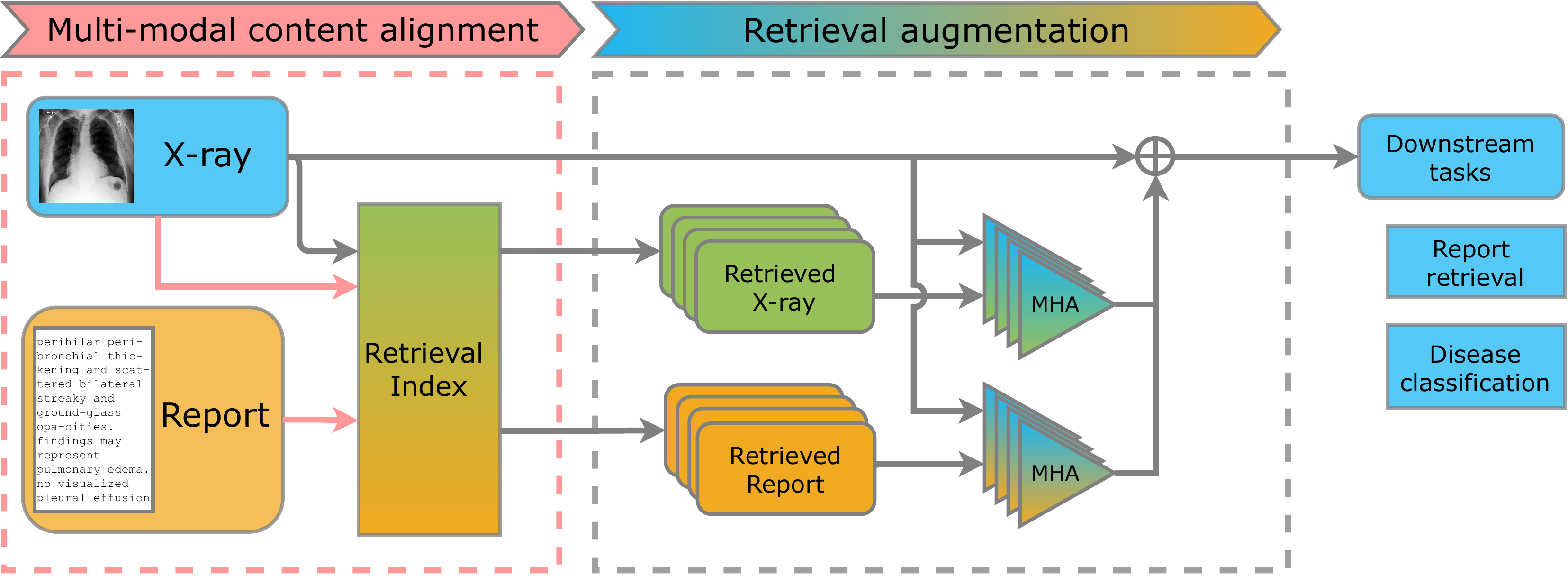}
    \caption{Architecture overview of X-TRA.}
    \label{fig:genarch}
\end{figure}

Our method is composed of two separate parts (\autoref{fig:genarch}). The first part is the alignment of the two modalities and construction of the retrieval model. The second part uses the output of the retriever as a non-parametric component in (cross-modal) retrieval augmentation to enhance the downstream tasks. 

We consider a dataset $\Theta^{N}_{\{\X,\R\}}$ consisting of pairs containing an X-ray ($\X_i$) and radiology report ($\R_i$). To align these modalities we make use of the powerful CLIP vision-language aligner. Our objective is to minimize the distance between $\X$ and $\R$, to make cross-modal tasks possible. These aligned features will be used for retrieval augmentation to do multi-label classification and report retrieval as downstream tasks. 
\subsection{Stage I: Multi-modal content alignment}

We leverage the pre-trained features from CLIP for initial feature representations. However, there is a domain shift between the natural image data CLIP is trained on and medical images we want to use in our method. Medical images can be visually very similar, while holding drastically different information. Small localized markers can be indicators for disease. In natural images global representations are more decisive and thus more suitable for unsupervised contrastive alignment. Alignment in CLIP goes as follows~\cite{radford2021learning},

\begin{equation}
\mathcal{L}_{CLIP}=-\frac{1}{N} \sum_{z \in Z}\sum_{i=1}^N  \log \frac{\mathrm{e}^{\left(\operatorname{sim}\left(z_i^0, z_i^1\right) / \tau\right)}}{\sum_{j=1}^N \mathrm{e}^{ \left(\operatorname{sim}\left(z_i^0, z_j^1\right) / \tau\right)}}\;\; with \;Z = \{(\X,\R),(\R,\X)\}.  
\end{equation}

We need to overcome the obvious domain shift between medical images and the natural images on which CLIP is trained. Therefore, we require a more specific type of fine-tuning that is especially geared towards content-based extraction. We introduce the following loss, requiring a global class label for each dataset. With this fine-tuning step we are creating a supervised content-based alignment method with content classifier $C$: 

\begin{equation}\qquad\mathcal{L}_{ours} = -\frac{1}{N}\sum_{z \in Z}\sum_{i=1}^N y_i log_{e}(\widehat{C(z_i)})\qquad\;\;\; with \; Z = \{\X,\R,(\X,\R)\}.\end{equation}

This content based alignment loss should improve the alignment of detailed content-level details over the global visual appearance of the image. 
\subsubsection{Creating a retrieval index}
At retrieval time we need to retrieve images that have a high similarity with query images. To efficiently do so we make use of Facebook AI Similarity Search (FAISS) \cite{johnson2019billion}. This retrieval tool efficiently performs nearest-neighbour similarity search. After multi-modal alignment we encode our data to a FAISS index $I$ conditioned on the entire training dataset. We can construct indices that only retrieve images ($I^{\X}$), only reports ($I^{\R}$), or both ($I^{\X\R}$). 

Given a query $\Q_s$ in source modality $s$, we can obtain its $k$ neighbours of target modality $t$ through: \begin{equation}\N_{\mathbf{s\veryshortrightarrow t}}^{k} = I^{t}(\Q_s,k),\end{equation} 
this can be either $\X$, $\R$ or both. Once retrieval index $I$ is trained based on the newly aligned training dataset we can consider the retriever as a non-parametric component which retrieves information from a fixed dataset in the subsequent retrieval augmentation steps. Note that during testing time, a query from the test set will be used to retrieve neighbours from the training set.    
\subsection{Stage II: Retrieval Augmentation}

The purpose of retrieval augmentation is to effectively leverage similar representations to adopt a more informative representation of a given input. with our already trained retrieval index we retrieve similar representations.

To obtain a richer representation of $\X_i$, we retrieve intra- $\N_{\mathbf{\X\veryshortrightarrow \X}}^{k}$  and inter- modal neighbours $\N_{\mathbf{\X\veryshortrightarrow \R}}^{k}$ from $I^{\X}$ and $I^{\R}$ respectively. To integrate the retrieved neighbouring samples, we can use various fusion methods~\cite{priyasad2021memory}. The simplest one is concatenation: $(\X_i, \N_{\mathbf{\X\veryshortrightarrow \X}}^{k},\N_{\mathbf{\X\veryshortrightarrow \R}}^{k})$. A more suitable method is multi-head attention (MHA) which is able to capture the long range dependencies between the original image and the retrieved information~\cite{vaswani2017attention}:
\begin{equation}
    \X^{TRA}_i = (\X_i, \mha(\N_{\mathbf{\X\veryshortrightarrow \X}}^{k},\X_i),\mha(\N_{\mathbf{\X\veryshortrightarrow \R}}^{k},\X_i)).
\end{equation}

\subsection{Downstream tasks}

We are tackling two common tasks in chest X-ray analysis. These are multi-label disease classification and report retrieval. For this last task our objective is simply to show how well a retriever can perform on the task of report generation. We measure the performance by comparing task performance of $\X^{TRA}$ in comparison to $\X$. 

A useful property of our retrieval index would be usability of an already trained model across datasets. Three clinically relevant scenarios for this are: From scratch training on the new dataset, frozen usage of the trained retrieval model and fine-tuning of the existing retrieval model with another image-report dataset.

\subsection{Datasets}
The primary dataset to which our method is applied is \textbf{MIMIC-CXR} \textit{(200k image-report pairs)}~\cite{johnson2019mimic}. Disease labels for each pair are extracted from the report through a rule-based extraction method~\cite{irvin2019chexpert}. To evaluate the versatility and cross-domain capabilities of our method, we use the small \textbf{openI} \textit{(4k image-report pairs)}~\cite{openi} and image-only \textbf{CheXpert} \textit{(200k images)}~\cite{irvin2019chexpert} datasets. Official train-test splits are used.  
\subsection{Experimental setup}

As pre-processing step, the X-ray images are normalized and standardized by rescaling with center-cropping to scale $256\times256$, from which images of size $224\times224$ are sampled. The maximum number of tokens for representing radiology reports in the text encoder is set to $256$. Three different VL models are used as encoders. At first a CNN-BERT model, composed of a DenseNet121 image encoder and a ClinicalBERT~\cite{huang2019clinicalbert} text encoder. This combination showed to worked well in previous multi-modal chest X-ray works. Given the strong performance of large vision-language models we also use CLIP (ViT-32 image encoder and text encoder)~\cite{radford2021learning} and its medically fine-tuned equivalent PubMedCLIP~\cite{EslamiDeMeloMeinel2021CLIPMedical}. This model is fine-tuned using the Radiology Objects in COntext (ROCO) dataset~\cite{pelka2018radiology}. 

Multi-modal alignment is implemented as a single pass through a two-layer ReLu activated MLP, with dimension $z_{enc}$, a dropout rate of $0.5$, and layer normalization. $z_{enc}$ is the output dimension of the encoder. We implement $C$ as a three layer classifier head with dimensions $\{z_{enc},256,14\}$. During retrieval we make use of $k=10$ retrieved neighbours. 
To prevent overfitting, early stopping with a tolerance of 3 is applied to all training operations.

\section{Results}

\subsection{Cross-modal Retrieval}
We are comparing the performance of our retrieval method against previous methods in \autoref{tab:retrieval_main} in terms of class-based mean average precision~(mAP). Due to the powerful alignment of CLIP and tailor made fine-tuning we are outperforming all existing retrieval approaches for radiology images and/or reports by a large margin. The performance difference with similarly fine-tuned encoder-decoder combination DenseNet121 and ClinicalBERT further underwrites the power of CLIP in building a strong retrieval method, specifically on cross-domain retrieval. Interestingly, we observe that PubMedCLIP is not outperforming CLIP. This can be explained by a domain shift between MIMIC-CXR and ROCO, together with the ability of CLIP to generalize well out-of-domain~\cite{radford2021learning}.
In our downstream tasks image-based retrieval is most important, which is performing similar on inter- and intra-modal retrieval tasks. 
\begin{table}[!t]
\centering
\caption{Class-based retrieval performance (source $\rightarrow$ target) for images ($\X$) and reports ($\R$) in terms of mAP on MIMIC-CXR on our content alignment method, compared against other methods.}
\label{tab:retrieval_main}
\resizebox{.68\textwidth}{!}{%
\begin{tabular}{llllcccccccccccccccc}
                      &&&                   & \rot{No Finding} & \rot{Enl. Cardiomed.} & \rot{Cardiomegaly} & \rot{Lung Opacity} & \rot{Lung Lesion} & \rot{Edema} & \rot{Consolidation} & \rot{Pneumonia} & \rot{Atelectasis} & \rot{Pneumothorax} & \rot{Pleural Effusion} & \rot{Pleural other} & \rot{Fracture} & \rot{Support Devices} & wAvg  & Avg   \\\cmidrule(){1-4}\cmidrule(rl){5-18}\cmidrule(rl){19-19}\cmidrule(l){20-20}
Yu et al.~\cite{yu2021multimodal}&\multirow{5}{*}{$\X$}&\multirow{5}{*}{$\rightarrow$}&\multirow{5}{*}{$\X$}&-&\textBF{.65}&.75&.72&.43&.80&.73&.60&.76&.76&.85&.43&.16&.86&-&.680\\
CLIP ($\mathcal{L}_{CLIP}$)&&&&  .71 & .52 & .74 & .78 & .39 & .79 & .39 & .40 & .76 & .42 & .67 & .44 & .43 & .64 & .578 & .761 \\
CNN+BERT    &  &  &  & .87 & .63 & .88 & \textBF{.90}  & .49 & .90  & \textBF{.57} & .60  & .85 & \textBF{.85} & .83 & .29 & .47 & .82 & .678 & .769 \\
PubmedCLIP  &  &  &  & \textBF{.90}  & .63 & .82 & .83 & .39 & .86 & .45 & \textBF{.63} & .87 & .53 & \textBF{.90}  & .48 & .51 & .79 & .685 & .795 \\
CLIP        &  &  &  & .84 & .62 & \textBF{.89} & .89 & \textBF{.56} & \textBF{.91} & .55 & .59 & \textBF{.89} & .60  & .86 & \textBF{.49} & \textBF{.57} & \textBF{.84} & \textBF{.713} & \textBF{.840}  \\\cmidrule(){1-4}\cmidrule(rl){5-18}\cmidrule(rl){19-19}\cmidrule(l){20-20}
Zhang et al.~\cite{zhang2022category}&\multirow{5}{*}{$\X$}&\multirow{5}{*}{$\rightarrow$}&\multirow{5}{*}{$\R$}&-&-&-&-&-&-&-&-&-&-&-&-&-&-&-&.498\\
CLIP ($\mathcal{L}_{CLIP}$) &  &  &  & .50 & .60 & .73 & .81 & .53 & .70 & .73 & .87 & .85 & .59 & .78 & .55 & .31 & .77 & .666 & .762 \\
CNN+BERT    &  &  &  & .61 & \textBF{.74} & .89 & \textBF{.95} & .45 & .69 & .76 & .82 & .71 & \textBF{.71} & .77 & .32 & .67 & \textBF{.84} & .713 & .756 \\
PubmedCLIP  &  &  &  & \textBF{.74} & .65 & .67 & .62 & .38 & .70  & .13 & .76 & .72 & .51 & .83 & .51 & \textBF{.90}  & .60  & .623 & .728 \\
CLIP        &  &  &  & .64 & .71 & \textBF{.91} & .92 & \textBF{.73} & \textBF{.87} & \textBF{.89} & \textBF{.94} & \textBF{.94} & .67 & \textBF{.95} & \textBF{.61} & .48 & \textBF{.84} & \textBF{.793} & \textBF{.793} \\\cmidrule(){1-4}\cmidrule(rl){5-18}\cmidrule(rl){19-19}\cmidrule(l){20-20}
CLIP ($\mathcal{L}_{CLIP}$) &\multirow{4}{*}{$\X$}&\multirow{4}{*}{$\rightarrow$}&\multirow{4}{*}{$\X\R$}& .76 & .66 & .81 & .88 & .61 & .73 & \textBF{.67} & .53 & .84 & .54 & .79 & .57 & .74 & .70 & .679 & .739 \\
CNN+BERT    &  &  &  & .85 & .85 & .76 & .75 & .51 & .83 & .64 & \textBF{.66} & .82 & .58 & \textBF{.95} & .53 & .62 & \textBF{.84} & .728 & .766 \\
PubmedCLIP  &  &  &  & \textBF{.90}  & .77 & .71 & .89 & \textBF{.81} & \textBF{.86} & .57 & .44 & .81 & .59 & .93 & .65 & .64 & .82 & .742 & .824 \\
CLIP        &  &  &  & .85 & \textBF{.86} & \textBF{.91} & \textBF{.90}  & .68 & .84 & .54 & \textBF{.66} & \textBF{.90}  & \textBF{.64} & .87 & \textBF{.68} & \textBF{.78} & .81 & \textBF{.780} & \textBF{.857} \\\cmidrule(){1-4}\cmidrule(rl){5-18}\cmidrule(rl){19-19}\cmidrule(l){20-20}
Zhang et al.~\cite{zhang2022category}&\multirow{5}{*}{$\R$}&\multirow{5}{*}{$\rightarrow$}&\multirow{5}{*}{$\X$}&-&-&-&-&-&-&-&-&-&-&-&-&-&-&-&.485\\
CLIP ($\mathcal{L}_{CLIP}$) &  &  &  & .62 & .52 & .93 & .88 & .50 & .60 & .29 & .44 & .75 & .54 & .85 & .50 & .36 & .71 & .606 & .723 \\
CNN+BERT    &  &  &  & \textBF{.77} & .54 & .73 & .91 & .52 & \textBF{.83} & .39 & \textBF{.87} & .77 & .63 & .74 & .23 & \textBF{.61} & .73 & .662 & .735 \\
PubmedCLIP  &  &  &  & .75 & \textBF{.65} & .92 & \textBF{.99} & .23 & .79 & .21 & .51 & .59 & \textBF{.72} & .81 & .56 & .43 & .67 & .645 & .720  \\
CLIP        &  &  &  & .63 & .62 & \textBF{.96} & .94 & \textBF{.62} & .69 & \textBF{.47} & .61 & \textBF{.85} & .69 & \textBF{.91} & \textBF{.57} & .46 & \textBF{.82} & \textBF{.703} & \textBF{.779} \\\cmidrule(){1-4}\cmidrule(rl){5-18}\cmidrule(rl){19-19}\cmidrule(l){20-20}
CLIP ($\mathcal{L}_{CLIP}$) &\multirow{4}{*}{$\R$}&\multirow{4}{*}{$\rightarrow$}&\multirow{4}{*}{$\R$}& .77 & .88 & .86 & .92 & .59 & .75 & .67 & .70 & .87 & .70 & .93 & .54 & .28 & .77 & .731 & .852  \\
CNN+BERT    &  &  &  & .83 & .63 & .86 & \textBF{.98} & .60  & .84 & .68 & .66 & .88 & .64 & .96 & .47 & \textBF{.54} & .80  & .741 & .843 \\
PubmedCLIP  &  &  &  & \textBF{.99} & .75 & \textBF{.90}  & \textBF{.98} & .67 & .84 & \textBF{.83} & .60  & \textBF{.95} & .82 & \textBF{.98} & .38 & .28 & \textBF{.84} & .772 & .887 \\
CLIP        &  &  &  & .93 & \textBF{.93} & .87 & .96 & \textBF{.73} & \textBF{.94} & .77 & \textBF{.79} & .87 & \textBF{.85} & .95 & \textBF{.55} & .42 & \textBF{.84} & \textBF{.814} & \textBF{.895} \\\cmidrule(){1-4}\cmidrule(rl){5-18}\cmidrule(rl){19-19}\cmidrule(l){20-20}
CLIP ($\mathcal{L}_{CLIP}$) &\multirow{4}{*}{$\R$}&\multirow{4}{*}{$\rightarrow$}&\multirow{4}{*}{$\X\R$}& .90 & .77 & .80 & .87 & \textBF{.77} & .72 & .61 & .74 & .86 & .77 & .69 & .31 & .28 & .80 & .707 & .828 \\
CNN+BERT    &  &  &  & .74 & .49 & .91 & .98 & .42 & .77 & .68 & .79 & .87 & .64 & .78 & .26 & \textBF{.61} & \textBF{.84} & .734 & .836 \\
PubmedCLIP  &  &  &  & \textBF{.92} & .81 & .94 & \textBF{.99} & .68 & \textBF{.86} & \textBF{.93} & \textBF{.82} & \textBF{.99} & .76 & .79 & .30  & .27 & \textBF{.84} & .793 & .903 \\
CLIP        &  &  &  & .91 & \textBF{.91} & \textBF{.96} & .97 & .76 & .84 & .76 & .77 & .92 & \textBF{.96} & \textBF{.84} & \textBF{.46} & .37 & .80  & \textBF{.803} & \textBF{.909}\\
                      \bottomrule
\end{tabular}%
}
\end{table}


\subsection{Multi-label disease classification}

Disease classification results in terms of AUC in \autoref{tab:main_cls} show that retrieval augmentation gives a clear improvement across different disease classes. It is interesting to see that we find a positive, albeit weak, correlation (R$\approx$0.60) between the increase in class AUC performance and retrieval mAP. Moreover, the performance gain from retrieval augmentation ($0.80\rightarrow0.85$) is similar to additional training with synthetic diffusion-generated X-rays ($0.80\rightarrow0.84$)~\cite{chambon2022roentgen}. The benefit of our method is that the supplemented information originates from the trusted dataset itself and is not synthetically generated. 

\begin{table}[!t]
\centering
\caption{Chest X-ray classification on MIMIC-CXR with and without retrieval augmentation. The results show the beneficial effect of retrieval augmentation on classification performance.}
\label{tab:main_cls}
\resizebox{0.73\textwidth}{!}{%
\begin{tabular}{lccccccccccccccccc}
                      &  \rot{X-TRA}                 & \rot{No Finding} & \rot{Enl. Cardiomed.} & \rot{Cardiomegaly} & \rot{Lung Opacity} & \rot{Lung Lesion} & \rot{Edema} & \rot{Consolidation} & \rot{Pneumonia} & \rot{Atelectasis} & \rot{Pneumothorax} & \rot{Pleural Effusion} & \rot{Pleural other} & \rot{Fracture} & \rot{Support Devices} & wAvg  & Avg   \\\cmidrule(r){1-2}\cmidrule(rl){3-16}\cmidrule(rl){17-17}\cmidrule(rl){18-18}

CNN+BERT&\xmark&.81 & .63 & .73 & .67 & .62 & .83  & .69 & .59 & .68 & .75  & .83 & .70  & .58 & .84 & .71 & .79 \\\rowcolor{gray!20}
&\checkmark&.81 & .74 & .75 & .69 & .63 & .81  & .72 & .63 & .75 & .75  & .83 & .69  & .63 & .85 & .73 & .82 \\
&$\Delta$&\cellcolor{red!0.0}.00&\cellcolor{red!100.0}.11&\cellcolor{red!20.0}.02&\cellcolor{red!20.0}.02&\cellcolor{red!10.0}.01&\cellcolor{red!20.0}.02&\cellcolor{red!30.0}.03&\cellcolor{red!40.0}.04&\cellcolor{red!70.0}.07&\cellcolor{red!0.0}.00&\cellcolor{red!0.0}.00&\cellcolor{yellow!10.0}-.01&\cellcolor{red!50.0}.05&\cellcolor{red!10.0}.01&\cellcolor{red!20.0}.02&\cellcolor{red!30.0}.03\\\cmidrule(r){1-2}\cmidrule(rl){3-16}\cmidrule(rl){17-17}\cmidrule(rl){18-18}

PubmedCLIP&\xmark&.78 & .65 & .72 & .66 & .61 & .82  & .70 & .61 & .73 & .76  & .81 & .62  & .54 & .84 & .70 & .78 \\\rowcolor{gray!20}
&\checkmark&.84 & .76 & .78 & .69 & .64 & .83  & .73 & .64 & .76 & .75  & .82 & .75  & .67 & .85 & .75 & .83 \\
&$\Delta$&\cellcolor{red!60.0}.06&\cellcolor{red!100.0}.11&\cellcolor{red!60.0}.06&\cellcolor{red!30.0}.03&\cellcolor{red!30.0}.03&\cellcolor{red!10.0}.01&\cellcolor{red!30.0}.03&\cellcolor{red!30.0}.03&\cellcolor{red!30.0}.03&\cellcolor{yellow!10.0}-.01&\cellcolor{red!10.0}.01&\cellcolor{red!100.0}.13 &\cellcolor{red!100.0}.13&\cellcolor{red!10.0}.01&\cellcolor{red!50.0}.05&\cellcolor{red!50.0}.05\\\cmidrule(r){1-2}\cmidrule(rl){3-16}\cmidrule(rl){17-17}\cmidrule(rl){18-18}

CLIP&\xmark&.77 & .65 & .71 & .67 & .62 & .85  & .73 & .61 & .72 & .75  & .80 & .59  & .51 & .83 & .70 & .80 \\\rowcolor{gray!20}
&\checkmark&.82 & .78 & .74 & .70 & .71 & .82  & .75 & .63 & .79 & .78  & .86 & .74  & .72 & .91 & .77 & .85 \\
&$\Delta$&\cellcolor{red!50.0}.05&\cellcolor{red!100.0}.13&\cellcolor{red!30.0}.03&\cellcolor{red!30.0}.03&\cellcolor{red!90.0}.09&\cellcolor{yellow!30.0}-.03&\cellcolor{red!20.0}.02&\cellcolor{red!20.0}.02&\cellcolor{red!70.0}.07&\cellcolor{red!30.0}.03&\cellcolor{red!60.0}.06&\cellcolor{red!100.0}.15 &\cellcolor{red!100.0}.21&\cellcolor{red!80.0}.08&\cellcolor{red!70.0}.07&\cellcolor{red!50.0}.05\\
\bottomrule
\end{tabular}%
}
\end{table}

\subsection{Report generation}
In retrieval augmented report retrieval we show interesting performance on the report generation metrics compared to a selection of previous methods. While it should not be expected that simple retrieval outperforms dedicated report generation methods we are able to provide a result that can be considered competitive (\autoref{tab:reportgen}). On the METEOR and ROUGE metric we are even outperforming most existing methods. The metrics reflect that the strength of report retrieval is indeed in the global representation of the report. Our retriever is fine-tuned to retrieve samples with equivalent label spaces, hence good results on metrics that reward global similarity. An interesting outlook is the application of this method in a dedicated report generation framework which could boost performance further.  


\begin{table}[!t]
\centering
\caption{Chest X-ray report retrieval  on MIMIC-CXR with and without X-TRA retrieval augmentation. compared to dedicated report generation methods.}
\label{tab:reportgen}
\resizebox{\textwidth}{!}{%
\begin{tabular}{lllllllll}
\toprule
&&  BLEU-1 & BLEU-2 & BLEU-3 & BLEU-4 & ROUGE-L & METEOR&BERTScore\\\cmidrule(rl){3-3}\cmidrule(rl){4-4}\cmidrule(rl){5-5}\cmidrule(rl){6-6}\cmidrule(rl){7-7}\cmidrule(rl){8-8}\cmidrule(rl){9-9}
\multirow{4}{*}{\begin{tabular}[c]{@{}l@{}}Report\\ generation\end{tabular}}&Pino \textit{et al.}~\cite{pino2021clinically}&-&-&-&.094&.185&-&-\\
&Wang \textit{et al.}~\cite{wang2022cross}&.344&.215&.146&.105&.279&.138&-\\
&Yang \textit{et al.}~\cite{yang2021writing}&.438&.297&.216&.164&.332&-&-\\
&Li \textit{et al.}~\cite{li2022self}&\textBF{.467}&\textBF{.334}&\textBF{.261}&\textBF{.215}&\textBF{.415}&.201&-\\\midrule
\multirow{5}{*}{\begin{tabular}[c]{@{}l@{}}Report\\ retrieval\end{tabular}}&Chambon \textit{et al.}~\cite{chambon2022roentgen}&-&-&-&-&-&-&.432\\
&Yang \textit{et al.}~\cite{yang2021writing}&.306&.179&.116&.076&.232&-&-\\
&CNN+BERT&.268~($\uparrow$.025)& .193~($\uparrow$.064)& .106~($\uparrow$.036) & .072~($\uparrow$.029) & .288~($\uparrow$.042)& .248~($\uparrow$.027)&.572($\uparrow$.17)\\

&PubmedCLIP&.308~($\uparrow$.031) & .206~($\uparrow$.021) & .111~($\uparrow$.021) & .074~($\uparrow$.006) & .330~($\uparrow$.022) & .286~($\uparrow$.025)&.610($\uparrow$.29)\\

&CLIP&.318~($\uparrow$.041) & .226~($\uparrow$.041) & .121~($\uparrow$.024) & .085~($\uparrow$.023) & .339~($\uparrow$.044) & \textBF{.296~($\uparrow$.055)}&\textBF{.617($\uparrow$.31)}\\

\bottomrule
\end{tabular}%
}
\end{table}

\subsection{Cross-dataset}
By evaluating the cross-dataset scenarios (\autoref{tab:cross_dataset}) with the CheXpert and openI datasets we can conclude that transferability to images from other domains is limited. However we do see that if retrieval augmentation is not useful, it can be ignored by the model and will not be detrimental for performance. The domain shift between different chest X-rays is a remaining problem \cite{pooch2019can}. Currently the most practical solution for this problem is the addition of a fine-tuning step.  

Cross-domain results on open-I show that learning across modalities is possible with fine-tuning. When adding the openI dataset to the existing retrieval index, we can integrate the existing index with this new dataset. We can see that X-TRA benefits openI in this setting. In the updated retrieval index $23\%$ of the retrieved information originates from openI and  $77\%$ from MIMIC-CXR.
\begin{table}[!t]
\centering
\caption{Cross-domain result on downstream tasks: Report retrieval (RR) and multi-label classification (MLC) with and without X-TRA.} 
\label{tab:cross_dataset}
\resizebox{.85\textwidth}{!}{%
\begin{NiceTabular}{lrcccccc}
\toprule

&Retrieval source$\rightarrow$&\multicolumn{2}{c}{Target}&\multicolumn{2}{c}{MIMIC-CXR}&\multicolumn{2}{c}{MIMIC-CXR}\\\cmidrule(rl){3-4}\cmidrule(rl){5-6}\cmidrule(rl){7-8}
\multirow{2}{*}{\begin{tabular}[c]{@{}l@{}}Target\\ $\downarrow$ dataset\end{tabular}}&Setting$\rightarrow$&\multicolumn{2}{c}{From scratch}&\multicolumn{2}{c}{Frozen}&\multicolumn{2}{c}{Finetuning}\\

                      &&  RR&MLC&  RR&MLC&  RR&MLC\\\cmidrule(rl){2-2}\cmidrule(rl){3-4}\cmidrule(rl){5-6}\cmidrule(rl){7-8}
CheXpert&CNN+BERT&-&-&-&.81($\downarrow$.01)&-&-\\
&PubmedCLIP&-&-&-&\textBF{.82}($\uparrow$.01)&-&-\\
&CLIP&-&-&-&.81(\phantom{$\uparrow$}.00)&-&-\\\midrule

OpenI&CNN+BERT&.31($\uparrow$.05)&.88($\uparrow$.01)&.31($\uparrow$.03)&.87($\uparrow$.01)&.33($\uparrow$.05)&.90($\uparrow$.05)\\

&PubmedCLIP&.26($\uparrow$.05)&.86($\downarrow$.01)&.34($\uparrow$.04)&.89($\uparrow$.03)&.38($\uparrow$.05)&.91($\uparrow$.05)\\

&CLIP&\textBF{.29}($\uparrow$.04)&\textBF{.90}($\uparrow$.04)&\textBF{.35}($\uparrow$.02)&\textBF{.90}($\uparrow$.02)&\textBF{.38}($\uparrow$.07)&\textBF{.93}($\uparrow$.05)\\

                      \bottomrule
\end{NiceTabular}%
}
\end{table}
\begin{figure}[!b]
    \centering
    \begin{subfigure}[b]{0.90\linewidth}
    \includegraphics[width=\linewidth]{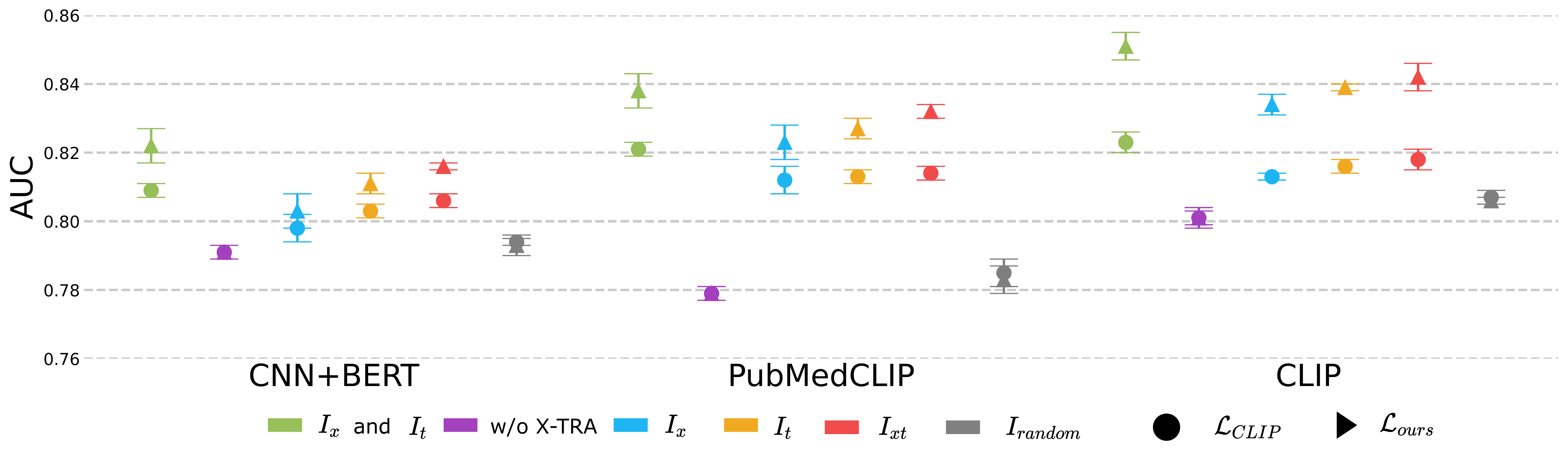}
    \caption{}
    \label{fig:ab2} 
    \end{subfigure}
    \begin{subfigure}[b]{0.90\linewidth}
    \includegraphics[width=\linewidth]{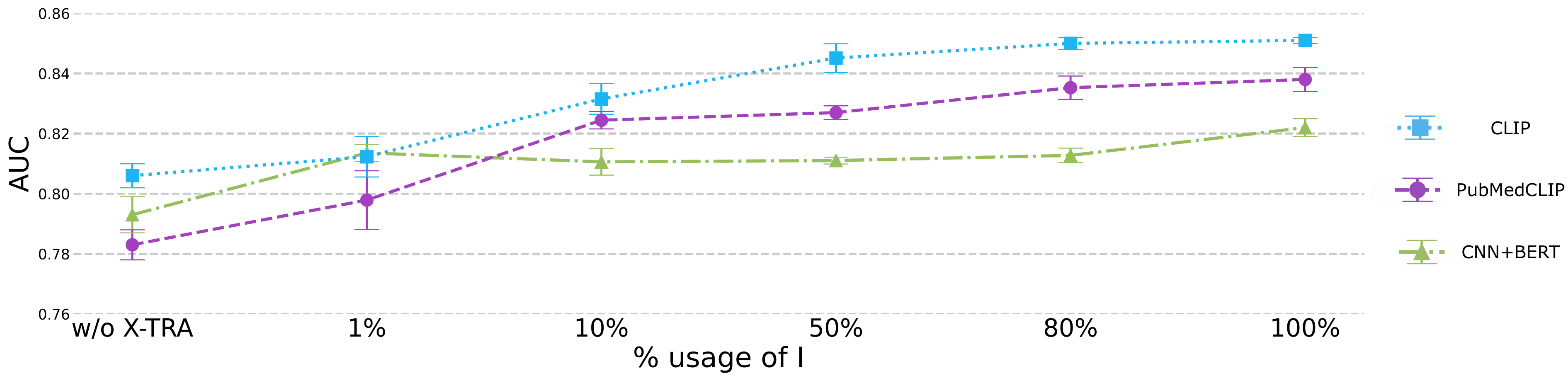}
    \caption{}
    \label{fig:ab1} 
    \end{subfigure}
    \caption{Ablation studies on X-TRA on disease classification, for five different random seeds, with (a) different compositions of the retrieval index for $\mathcal{L}_{CLIP}$ and $\mathcal{L}_{ours}$ and (b) partial usage of the retrieval index.}
    \label{fig:abl}
\end{figure}
\subsection{Ablation studies}
We study the effect of the components in our retrieval augmentation method in \autoref{fig:abl}. Specifically we look at the influence of each component in content- and CLIP based alignment. Interestingly, the composition of data modalities in retrieval augmentation does not have a big effect, since the retriever has similar results in inter- and intra-modal retrieval. In case randomly selected data is used instead of retrieved information, we achieve comparable results compared  to our method without X-TRA. This is in accordance with cross-modal results, showing that if X-TRA supplemented information is not useful, it can be ignored. Using a partial retrieval index we can conclude that X-TRA can be useful with a small retrieval index, however performance reaches optimal levels when $N>100k$.

\subsection{Insight and limitations}
\begin{figure}[!t]
    \centering
    \includegraphics[width=0.9\linewidth]{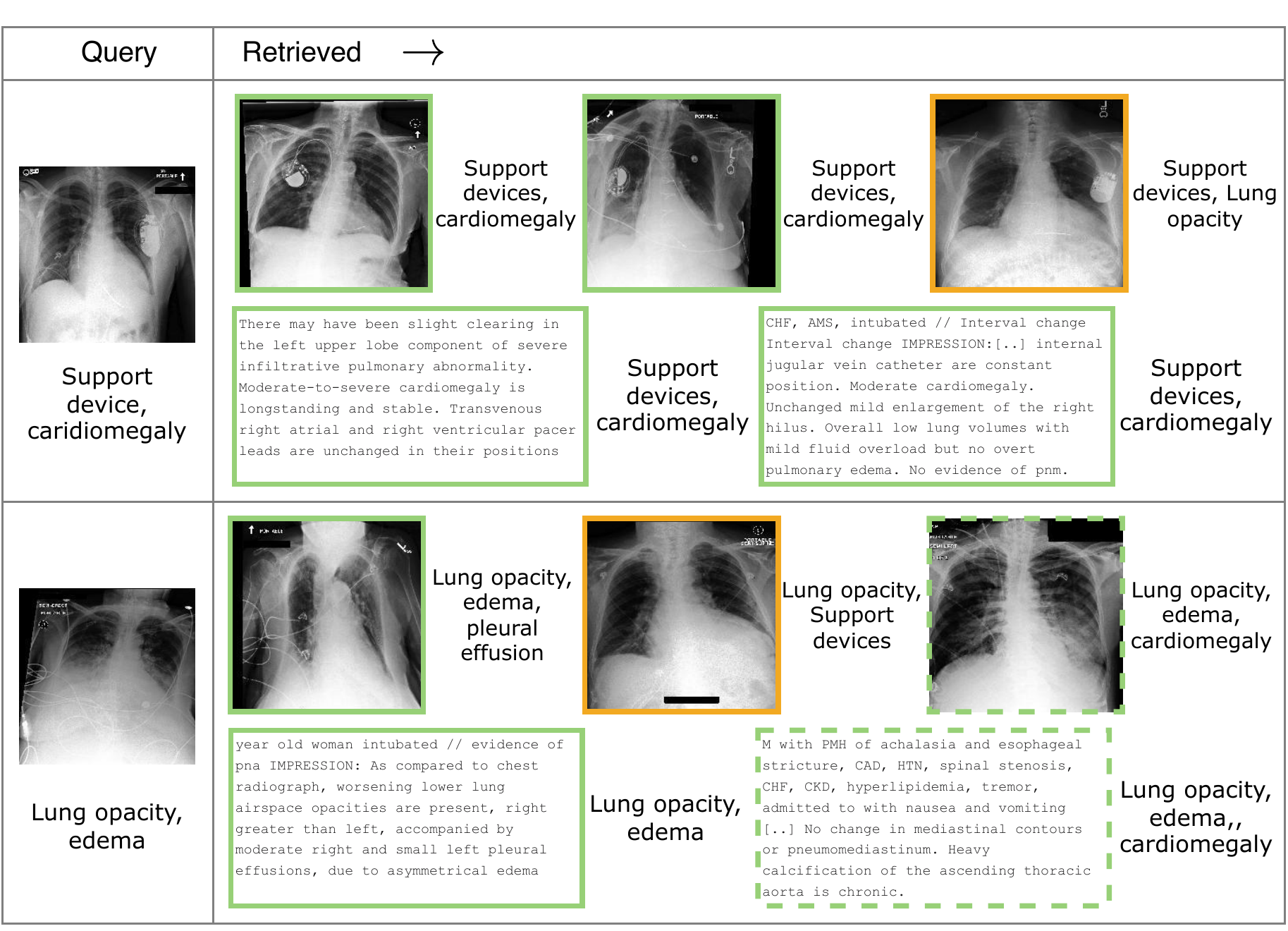}
    \caption{Examples of image-image and image-text retrieval including disease class labels. A green outline means a correct retrieval, orange or dashed means a missed or extra disease label respectively.}
    \label{fig:q}
\end{figure}

Qualitative results from our retrieval method for 2 different query images is shown in \autoref{fig:q}. We retrieve from the image index and report index. The retrieved images match well in terms of labels attributed to them, showing that our fine-tuning is preventing the retrieval of images that are only globally similar. 

Fine-tuning of the entire CLIP model to domain-specific data is an interesting prospective. Potentially this can further improve the performance of our retrieval model. However, as we have shown in this paper regarding the performance of CLIP against PubMedCLIP, the loss of generalization can also be detrimental. In future studies this an promising avenue to explore. 
\section{Conclusion}

In this work we present X-TRA, a simple yet effective method to improve multiple tasks on radiology images. Our method is composed of a content alignment and a retrieval augmentation step. With a new label-based alignment loss we are able to leverage pre-trained CLIP features to create a powerful cross-modal retrieval model. The general CLIP model appears to be more useful for our retrieval model than the slightly out-of-domain medically fine-tuned PubMedCLIP. We use this retrieval model to improve chest X-ray analysis through retrieval augmentation. With this we are adding an enrichment and regularization component that improves both multi-label disease classification and report retrieval by up to over $5\%$. On this last task we are even showing to be competitive with dedicated report retrieval methods. It opens up possibilities for retrieval augmentation as a generic tool in medical imaging.

\bibliographystyle{splncs04}
\bibliography{sources}
\end{document}